\documentclass[conference]{IEEEtran}
\IEEEoverridecommandlockouts

\usepackage{cite}
\usepackage{amsmath,amssymb,amsfonts}
\usepackage{algorithmic}
\usepackage{graphicx}
\usepackage{textcomp}
\usepackage{xcolor}
\usepackage{booktabs}
\usepackage{float}
\usepackage{makecell}

\def\BibTeX{{\rm B\kern-.05em{\sc i\kern-.025em b}\kern-.08em
    T\kern-.1667em\lower.7ex\hbox{E}\kern-.125emX}}
\begin{document}

\title{Empowering Epidemic Response: The Role of Reinforcement Learning in Infectious Disease Control\thanks{Accepted at the 6th International Workshop on AI for Social Good in the Connected World, held in conjunction with the 24th IEEE/WIC International Conference on Web Intelligence and Intelligent Agent Technology (WI-IAT 2025).}}

\author{\IEEEauthorblockN{Mutong Liu}
\IEEEauthorblockA{\textit{Department of Computer Science} \\
\textit{Hong Kong Baptist University}\\
Hong Kong SAR, China \\
csmtliu@comp.hkbu.edu.hk}
\and
\IEEEauthorblockN{Yang Liu}
\IEEEauthorblockA{\textit{Department of Computer Science} \\
\textit{Hong Kong Baptist University}\\
Hong Kong SAR, China \\
csygliu@comp.hkbu.edu.hk}
\and
\IEEEauthorblockN{Jiming Liu}
\IEEEauthorblockA{\textit{Department of Computer Science} \\
\textit{Hong Kong Baptist University}\\
Hong Kong SAR, China \\
jiming@comp.hkbu.edu.hk}
}
\maketitle

\begin{abstract}
Reinforcement learning (RL), owing to its adaptability to various dynamic systems in many real-world scenarios and the capability of maximizing long-term outcomes under different constraints, has been used in infectious disease control to optimize the intervention strategies for controlling infectious disease spread and responding to outbreaks in recent years.
The potential of RL for assisting public health sectors in preventing and controlling infectious diseases is gradually emerging and being explored by rapidly increasing publications relevant to COVID-19 and other infectious diseases.
However, few surveys exclusively discuss this topic, that is, the development and application of RL approaches for optimizing strategies of non-pharmaceutical and pharmaceutical interventions of public health.
Therefore, this paper aims to provide a concise review and discussion of the latest literature on how RL approaches have been used to assist in controlling the spread and outbreaks of infectious diseases, covering several critical topics addressing public health demands: resource allocation, balancing between lives and livelihoods, mixed policy of multiple interventions, and inter-regional coordinated control.
Finally, we conclude the paper with a discussion of several potential directions for future research.
\end{abstract}


\section{Introduction}

\begin{figure*}[!t]
    \centering
    \includegraphics[width=0.75\linewidth]{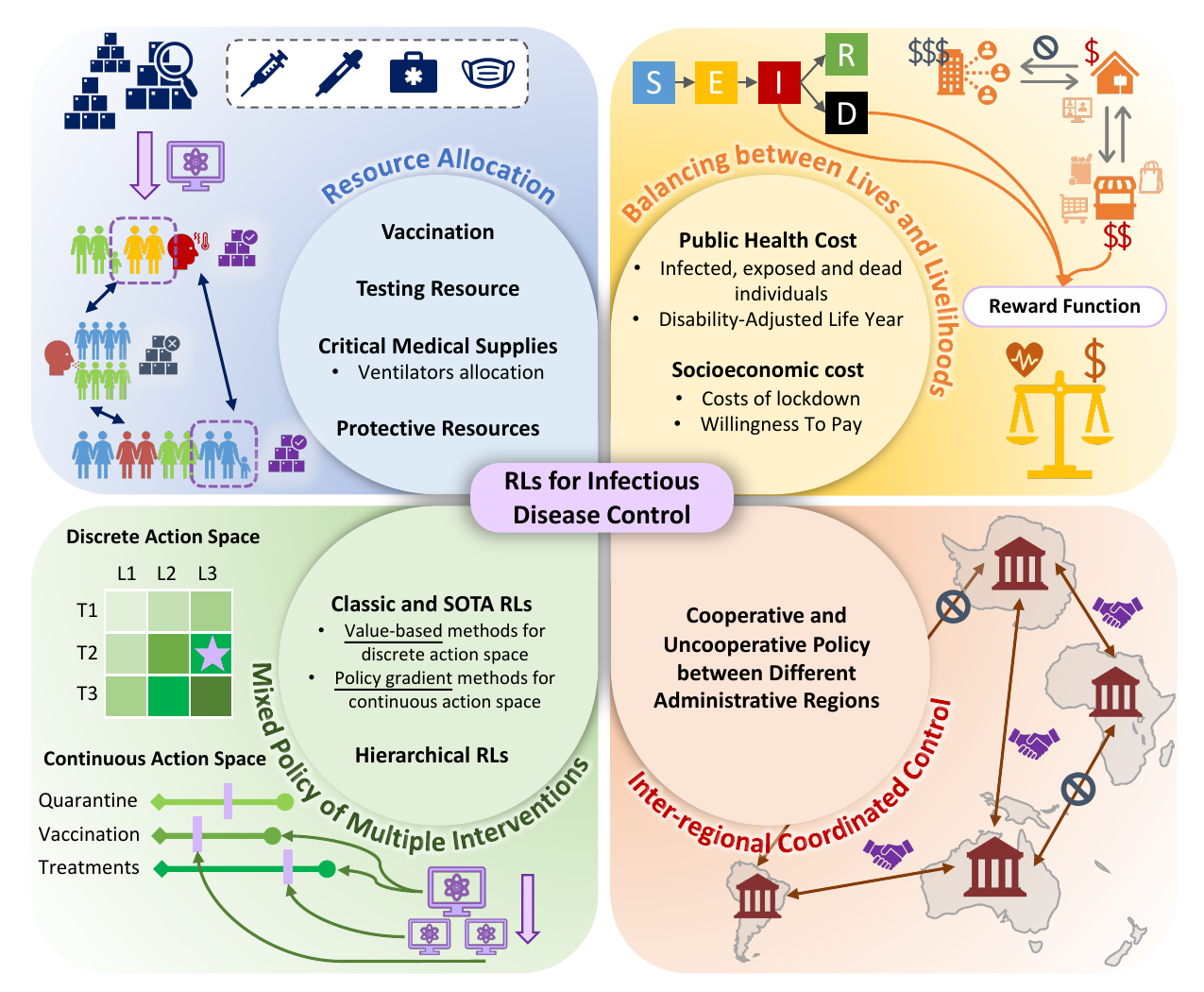}
    \caption{The taxonomy of reinforcement learning for epidemic control.}
    \label{fig:Structure}
\end{figure*}

Globally, the spread of infectious diseases, such as COVID-19, influenza, and HIV, poses persistent threats to population health, strains public health systems, and undermines the economic stability of society. During the disease propagation, public health authorities and relevant government sectors are tasked with deploying available interventions, including non-pharmaceutical interventions (NPIs) (e.g., social distancing and lockdowns) and pharmaceutical interventions (e.g., vaccination and medical treatment). However, due to the inherent complexity and stochasticity of infectious disease transmission driven by diverse factors \cite{heesterbeek2015modeling, kraemer2025artificial}, it is difficult to manually design control strategies solely based on expert knowledge or exhaustively select appropriate intervention policies through traditional simulation to effectively respond to rapidly evolving epidemic dynamics.

To tackle these decision-making challenges, Reinforcement Learning (RL) presents a powerful and applicable framework.
RL is able to address sequential decision-making problems under uncertain and complex environments, capable of adaptively learning optimal or near-optimal strategies through exploration in simulated environments.
Its success in diverse domains such as autonomous driving \cite{zhao2024survey}, robotic systems \cite{singh2022reinforcement}, and game theory \cite{silver2016mastering, silver2017mastering} underscores its potential for infectious disease control.
With RL's capacity to tackle critical adaptive optimization problems in infectious disease control, decision makers can determine \textbf{WHEN} and \textbf{HOW} to deploy \textbf{WHAT} interventions to reduce the overall and long-term disease risks while considering diverse constraints in the real world.

In existing literature, there are several surveys providing systematic introductions to applications of RL methods in the field of healthcare and public health across diverse topics \cite{yu2021reinforcement, weltz2022reinforcement, mishra2025disease}, e.g., disease diagnosis, dynamic treatment regimes, resource scheduling, drug discovery, etc.
However, few surveys exclusively discuss the up-to-date developments and applications of RL methods for optimizing intervention strategies from the perspective of practical topics concerned during public health implementation.

In this paper, we provide a concise review of the latest literature (published within the last five years) on how the RL approaches have been used to assist in controlling the spread and outbreaks of infectious diseases using various interventions.
Typically, an RL agent serving as the role of the public health authority or the government sector interacts with an environment, which is modeled by a deterministic or stochastic simulation of the disease transmission process, and gradually learns to derive appropriate intervention strategies, under various practical constraints.
Translating the general RL framework into practice for infectious disease control entails tackling several key challenges relevant to the actual needs of public health.
Accordingly, this survey organizes the latest literature by focusing on four pivotal topics with different problem definitions and formulations: (1) allocating resources under scarcity; (2) balancing public health risks and socioeconomic costs (i.e., lives and livelihoods); (3) optimizing mixed policies of multiple interventions; and (4) enabling inter-regional coordinated control. The taxonomy of our survey is illustrated in Fig. \ref{fig:Structure}.

The remainder of this paper is organized as follows. Section \ref{sec: Method} introduces the searching scope and methodology of our survey. Sections \ref{sec: Resource}, Section \ref{sec: Balancing}, \ref{sec: Multiple Interventions}, and \ref{sec: Inter-regional} present the core of this paper which reviews and categorizes RL applications in epidemic control on the four topics: resource allocation, balancing between lives and livelihoods, mixed policy of multiple interventions, inter-regional coordinated control, respectively. Section \ref{sec: Future} concludes the paper with the discussion of future directions.

\section{Systematic Literature Review Methodology}\label{sec: Method}
To catch the latest trend in the development and application of RLs on infectious disease control, we conducted a comprehensive literature search on the PubMed database with the search range from 2020 to July 8, 2025, with the search keywords including ("reinforcement learning") AND (("infectious disease" control) OR (epidemic control) OR ("infectious disease" intervention strategy)). In this step, we identified 49 publications by searching the PubMed database with the above-mentioned keywords and cutoff time.

By screening the title and abstract of each paper, we excluded survey papers, works focusing on environmental design and development, and papers with topics that do not fall within the range of optimizing intervention strategies for controlling infectious disease spread using RL methods. After checking the accessibility of the full text, two papers were excluded due to restricted access. We finally kept 19 papers reviewed and discussed in our survey.

\section{Resource Allocation}\label{sec: Resource}
Effective epidemic response requires the strategic allocation of limited resources to high-risk populations or regions to mitigate disease transmission \cite{emanuel2023shared}. Recently, a rising body of research employs RL methods to optimize allocation strategies by accounting for the long-term impact of interventions, thereby minimizing the overall risk of the epidemic. The summary of the different types of scarce resource allocation and adopted RL methods in terms of existing literature included in this survey is presented in Table \ref{tab: Resource allocation}.
Specifically, Dong et al. formulated the allocation of limited vaccination as the problem of identifying and removing a set of influencers in the virus transmission network at the individual level \cite{dong2024integrating}. The policy is designed as a sequential process learned by Q-learning, which is a classic value-based RL method, to select and remove one node until achieving the maximum number of nodes needed to be selected (e.g., 20\% of nodes).
Further introducing and modeling human mobility between urban regions from a macro perspective, Ling et al. utilized the Actor-Critic (AC) method with Proximal Policy Optimization (PPO), which is a type of policy gradient RL algorithm, to generate the vaccine distribution strategy at the meta-population level \cite{ling2024cooperating}. Given the total amount of daily vaccine supply, their algorithm allocates the vaccine to susceptible populations in different regions to minimize the increase in infected cases and deaths.
\begin{table}[!thbp]
    \centering
    \caption{Summary of the types of limited resource and utilized RL methods in the resource allocation.}
    \label{tab: Resource allocation}
    \scalebox{1}{
    \begin{tabular}{c|>{\centering\arraybackslash}m{4cm}|>{\centering\arraybackslash}m{2cm}}
      \toprule
      Refer. & Limited Resource Allocation & RL Methods \\
      \midrule
      \cite{dong2024integrating} 2024 & Vaccinating 20\% individuals at social network & Q-learning \\
      \midrule
      \cite{ling2024cooperating} 2024 & Distributing vaccines to different census block zones, given a fixed number of vaccinations & Actor-critic \\
      \midrule
      \cite{bastani2021efficient} 2021 & Conducting PCR testing for travelers at 40 entry points on Greek borders with 400 health workers, 32 laboratories across the country & Multi-armed bandit \\  
      \midrule
      \cite{bednarski2021collaborative} 2021 & Exchanging vacant ventilators based on the total number of initial ventilators of all states & Value iteration and Q-learning \\
      \midrule
      \cite{wang2020risk} 2020 & Distributing protective resources to a set of individuals at the social network & Q-learning \\
      \bottomrule
    \end{tabular}
    }
\end{table}

Optimal allocation of other types of scarce resources during the pandemic, such as testing resources, critical medical supplies, and protective resources, is also studied in existing works.
Bastani et al. aimed to allocate the limited testing resource by deploying a reinforcement learning system, nicknamed Eva, at the 40 entry points of the Greek borders during the COVID-19 pandemic \cite{bastani2021efficient}. Firstly, they classify the incoming travelers into different types based on the collected demographic features, and then, based on the different types of travelers, they formulate the testing allocation as the problem of multi-armed bandits. Specifically, each type of traveler can be recognized as an arm of bandits, and the goal is to maximize the number of infected asymptomatic travelers identified after assigning the Polymerase Chain Reaction (PCR) testing to the arms while balancing the exploitation for high-prevalence groups and exploration of the types with fewer samples.
Bednarski et al. applied two RL methods, i.e., value iteration and Q-learning, to optimize ventilator redistribution among US states to relieve resource shortage during the epidemic \cite{bednarski2021collaborative}. They assume that the initial number of ventilators in a state is equal to the number of beds in the COVID-19 intensive care unit (ICU). To share the idle ventilators with each other, their model mainly consists of the demand prediction, which utilizes the deep neural networks to estimate future state-wise ventilator demands with the prediction interval indicating the logistics delay, and the redistribution algorithm, which optimizes the state-to-state transfer matrix by punishing the resource shortage defined with the ratio of supply and demand.
Wang et al. studied the problem of using the graph theory method and Q-learning to allocate the limited protective resources to the highly influential individuals in the dynamic contact network to curb the disease transmission process \cite{wang2020risk}. By defining the weighted undirected graph between individuals with the data collected from the mobile devices, they formulate the problem of epidemic control as the Minimum Weight Vertex Cover (MWVC) problem from graph theory, and optimize the solution via Q-learning. By adaptively selecting a set of nodes in the graph, the algorithm aims to cover as many edge weights as possible while minimizing the number of selected nodes to achieve the optimal allocation of protective resources.

In summary, based on the various public health needs, works presented in this section formulate different paradigms based on various studied problems, such as the critical node group identification problem in the complex network \cite{dong2024integrating, wang2020risk} and the multi-armed bandit \cite{bastani2021efficient}, to address the allocation of scarce resources, providing inspiration for subsequent methodological and practical innovations in this topic.

\section{Balancing between Lives and Livelihoods}\label{sec: Balancing}
\begin{table*}[!tbp]
    \centering
    \caption{Summary of intervention types and value ranges, health risks concerned by public health, and socioeconomic effects caused by interventions, integrating to balance between lives and livelihoods. The arrows denote the goal of indicator optimization, where $\uparrow$ denotes as higher as better (maximizing) and $\downarrow$ denotes as lower as better (minimizing).}
    \label{tab: Balance}
    \scalebox{0.88}{
    \begin{tabular}{c|>{\centering\arraybackslash}m{4cm}|>{\centering\arraybackslash}m{4cm}|>{\centering\arraybackslash}m{5cm}}
      \toprule
      Refer. & Interventions & Public Health Effects & Socioeconomic Effects \\
      \midrule
      \cite{khadilkar2020optimising} 2020 & Region lockdown (Binary 0 or 1) & Number of infected population and deaths ($\downarrow$) & Total costs of each lockdown day ($\downarrow$) \\
      \midrule
      \cite{ohi2020exploring} 2020 & Movement restrictions (3-level percentage \{0, 25\%, 75\%\}) & Number of deaths ($\downarrow$) & Ratio of economic benefits brought by mobility ($\uparrow$) \\
      \midrule
      \cite{an2021dynamic} 2021 & Multi-modal inter-city travel ban (Binary 0 or 1) & Number of confirmed cases and deaths ($\downarrow$) & Total retail sales of consumer goods contributed by traffic turnover volume ($\uparrow$) \\
      \midrule
      \cite{song2022pandemic} 2022 & Allowed inter-community mobility (Percentage $[0, 100\%]$) & Number of hospitalized population ($\downarrow$) & Difference between the mobility demand and the actual mobility retention ($\downarrow$) \\
      \midrule
      \cite{beigi2021application} 2021 & Vaccination for susceptible population (Percentage $[0, 100\%]$) & Number of infected and exposed individuals ($\downarrow$) & Ratio of vaccinated people ($\downarrow$) \\
      \midrule
      \cite{nguyen2022general} 2022 & Social Distancing (SD) level (Decimal within $[0,\text{SD}_{max}]$) & Disability-Adjusted Life Year (DALY) losses averted by SD intervention ($\uparrow$) & Total amount of willingness to pay for each averted DALY ($\downarrow$) \\ 
      \bottomrule
    \end{tabular}
    }
\end{table*}

When considering how to deploy the NPIs, such as quarantining, closing schools, businesses, and workplaces, keeping social distancing, staying home if sick, wearing facial coverings, avoiding gatherings, and pharmaceutical interventions, such as vaccination and treatment, the effectiveness of those intervention strategies with different types and intensity on controlling epidemic spread is critical, which is relevant to the health and well-beings of the population, especially for the population more susceptible and with low immunity \cite{world2023managing}.
However, the strict strategy to prevent and contain the epidemic is not applicable persistently because it will lead to huge financial costs to governments and individual livelihoods \cite{krauss2022prevent}. For example, widespread quarantine and lockdown would suspend economic activities as well as restrict the freedom of individual mobility; taking tests and vaccinations to the whole population without distinction will also lead to numerous economic costs to governments and overwhelm the healthcare capacity.
Therefore, how to balance the health risk and economic cost, specifically in the complex and dynamic real-world environments, is a vital problem. A large amount of literature explores the solution to this problem using RL methods.
Typically, this problem can be formulated with the multi-objective RLs. The current methods mainly focus on the trade-off between the health cost of the population (e.g., the number of infections and the strain of the pandemic on the healthcare system) and the economic cost caused by the employed intervention (e.g., the economic cost of interventions and the impact of interventions on the economic activities).

Targeting diverse interventions, different types of evaluation for health risk and livelihood costs are designed in the existing literature. The summary of intervention types and value ranges, public health effects, and socioeconomic effects for each work in this section is summarized in Table \ref{tab: Balance}.
Aimed at the mobility restrictions, several works have developed various reward functions based on different assumptions for the impacts of movement on the economy.
For instance, Khadilkar et al. learned the optimal lockdown policies while balancing both health and economic costs via directly predefining the cost of each lockdown day. They constructed a disease transmission model with human mobility among 100 regions based on a classic epidemiological model with 6 compartments (susceptible, exposed, infected, asymptomatic carrier, dead, and recovered) \cite{khadilkar2020optimising}. They designed a reward function additively integrating the negative costs of each lockdown day (for each region, using a binary value, 1 or 0, to indicate lockdown or not), each infected person, and each death, and assigned each item with the specified weights 1.0, 1.0, and 2.5, respectively, to calculate the reward signal for the RL algorithm optimization.
Ohi et al. \cite{ohi2020exploring} and An et al. \cite{an2021dynamic} assume that the mobility will contribute to the economic promotion and calculate the economic benefit brought by the available mobility.
Specifically, Ohi et al. \cite{ohi2020exploring} defined three movement restriction levels and quantifying corresponding intensity: level 0, indicating all people can freely move as their usual patterns; level 1, indicating 25\% daily movement is cut to mimic the strictness of maintaining social distancing; and level 2, indicating 75\% daily movement is cut to mimic the scenario of a nationwide lockdown. Furthermore, they assume that the movement of each individual can contribute to the economic activity with a value within $[0.8,1]$, and the infectious and dead population does not contribute to the economy. Based on the assumption, the designed reward function is composed of two components: the economic benefits brought by the available individuals, where the threshold is constrained by the number of active cases, and the penalty of death, which is directly calculated by the number of deaths, to guide the RL model, i.e., Double Deep Q-Learning (DDQN), learns the daily movement restrictions.
An et al. considered the effects of ban strategies of inter-city travel of multi-modal transportation, i.e., airplane, high-speed train, coach, and car, on disease transmission and economic outcomes \cite{an2021dynamic}. Based on the constructed inter-city transportation networks, they aim to learn when to open and close the diverse transportation modes. They assume that the mobility will promote the regional consumption and thus utilize the Total Retail Sales of Consumer Goods (TRSCG) as the indicator of the economy in the reward function, and trade it off against the cost of epidemic patient treatment (i.e., confirmed cases) and deaths.
Song et al. also explored the effects of mobility-control policies by making the trade-off between the infection suppression and the mobility constraint strictness \cite{song2022pandemic}. Differently, they consider the effects of the difference between movement demands and actual mobility, led by mobility restriction. Based on the settings of mobility retention matrices of communities, they consider and evaluate three types of mobility-related control measures with different flexibility to those matrices: (1) city lockdown, (2) community quarantine, and (3) route management. To balance the requirement of curbing disease transmission and the effects of mobility-control policies, they define a reward function integrating the infection-spread cost, calculated by the number of hospitalized population, and the mobility-restriction cost, calculated by the difference between the mobility demand and the actual mobility retention.

Targeting the vaccination strategy, Beigi et al. applied the AC algorithm to optimize the vaccine allocation strategies aiming at the susceptible population of an epidemiological model at the population level by designing four types of objective functions \cite{beigi2021application}. Those objectives consider different compartments in the epidemic simulation and are formulated as reducing the weighted sum of (1) the infected individuals and the ratio of vaccinated people in terms of the susceptible population, (2) the infected individuals, exposed individuals, and the ratio of vaccinated people, (3) the quarantined exposed individuals plus three items in the second objective, (4) the fraction of quarantined susceptible individuals who vaccinated plus four items in the third objective.
Nguyen and Prokopenko aimed to learn the cost-effectiveness decision-making of Social Distancing (SD) by PPO \cite{nguyen2022general}. Given that they assume the individuals in the agent-based simulation environment are not homogeneous and do not consistently adhere to the SD measure by the government, they do not obtain an SD level and consistently apply it to all the individuals, but rather optimize the maximum level of the SD compliance via an RL agent.
To balance the health and economic cost, they define the Net Health Benefit (NHB), which is the difference between the health effects and the cost incurred by the intervention, as the reward signal for the RL agent.
Specifically, the health effect is quantified by the difference of Disability-Adjusted Life Year (DALY) without SD restriction and with SD restriction (i.e., averted DALY by intervention) instead of the number of infected or exposed individuals, like the above-mentioned literature; the cost of intervention is quantified by the total amount of Willingness To Pay (WTP) for each averted DALY by intervention (i.e., predefined financial expenditure of \$10K, \$50K, and \$100K per DALY).

The common essential of works in this section is designing multi-objective indicators, incorporating both health risk and economic cost, to provide the signals for the RL algorithm optimization. However, some of them still face the biases of reward shaping. Specifically, they adopt the plain weighted sum of life cost and livelihood cost to calculate an overall reward value \cite{khadilkar2020optimising, an2021dynamic}. The learned policy could overlook the valuable alternative balancing points in the pursuit of a single, aggregated reward signal, especially in the case of inappropriate selection of the weights and inaccurate quantification for health damage and economic consequences of intervention.

\section{Mixed Policy of Multiple Interventions}\label{sec: Multiple Interventions}
In previous sections, we discussed the optimization of the epidemic control policy for various single interventions under different problem backgrounds.
During the epidemic and pandemic, governments typically deploy multiple interventions simultaneously to effectively and rapidly curb the disease transmission \cite{brauner2021inferring}. However, the effects of combined control interventions vary with the rapidly evolving disease dynamics \cite{sharma2021understanding}. Furthermore, as the types of available interventions increase, the number of possible combinations becomes enormous. This makes the adaptive selection of the optimal or near-optimal policy very difficult.
Therefore, existing work explores RL methods to optimize the adaptive combination of several interventions during the disease evolution. The summary of related literature is summarized in Table \ref{tab: multi-interventions}.

\begin{table}[!tbp]
    \centering
    \caption{Summary of targeted diseases, multiple interventions, and adopted RL methods for each work in the mixed policy of multiple interventions.}
    \label{tab: multi-interventions}
    \scalebox{0.75}{
    \begin{tabular}{c|c|c|c}
      \toprule
      Refer. & Diseases & Interventions \& Action Space & RLs \\
      \midrule
      \makecell[c]{\cite{khatami2021reinforcement}\\2021} & HIV & \makecell[l]{Discrete domain: $2 \times 3 \times 2 \times 3$ \\ 1. Diagnostic rate for HETs \\ (Decrement of unaware population rate): $\{0\%, -2.5\%\}$ \\ 2. Retention-in-care rates for HETs \\ (Increment of ART proportion): \{0\%, 10\%, 20\%\} \\3. Diagnostic rate for MSM: same as 1 \\ 4. Retention-in-care rates for MSM: same as 2} & Q-learning \\
      \midrule
      \makecell[c]{\cite{kwak2021deep}\\2021} & COVID-19 & \makecell[l]{Discrete domain: $3 \times 3$ \\
      1. Domestic lockdown: \{L0, L1, L2\} \\ 2. Travel restrictions: \{T0, T1, T2\}} & D3QN \\
      \midrule
      \makecell[c]{\cite{ghazizadeh2024modeling}\\2024} & COVID-19 & \makecell[l]{Continuous domain for each: $[0,1]$ \\ 1. Quarantine of susceptible \\ 2. Vaccination \\ 3. Treatments} & DDPG \\
      \midrule
      \makecell[c]{\cite{kao2024dynamic}\\2024} & COVID-19 & \makecell[l]{Continuous domain: \\ 1. Moving distances: $[1, 5]$ \\ 2. Radius of screening area: $[0, 10]$} & A3C \\
      \midrule
      \makecell[c]{\cite{mannarini2022if}\\2022} & COVID-19 & \makecell[l]{Continuous domain for each: $[-1,1]$ \\ 1. Cancel public events \\ 2. Close public transport \\3. International travel controls \\ 4. Public information campaigns \\ 5. Gathering restrictions \\ 6. Internal movement restrictions \\ 7. School closing \\ 8. Stay at home requirements \\ 9. Workplace closing} & DDPG \\
      \midrule
      \makecell[c]{\cite{bushaj2023simulation}\\2023} & COVID-19 & \makecell[l]{Discrete domain: 9 actions\\1. Do nothing \\ 2. Testing, contact tracing, and quarantine \\ 3. Close schools and non-essential workplaces \\ 4. Mandatory mask \\ 5. Testing, contact tracing, quarantine, \\ close schools, and non-essential workplaces \\6. testing, contact tracing, quarantine, and \\ mandatory mask \\ 7. Vaccination \\8. Vaccination and mandatory mask \\ 9. Total lockdown} & DRL\\
      \midrule
      \makecell[c]{\cite{du2023hrl4ec}\\2023} & COVID-19 & \makecell[l]{Continuous and discrete domain: \\1. Mobility constraint: $[0, 100\%]$ \\ 2. Temporary medical resources: \{L, M, H\} \\ 3. Necessities supply: \{Non,L,M\}} & \makecell[c]{Hierarchical \\ PPO} \\
      \bottomrule
    \end{tabular}
    }
\end{table}

Several works applied classical value-based RL methods to find the optimal intervention combinations in the limited discrete state-action space.
For example, Khatami et al. applied Q-learning to optimize combinations of testing and retention-in-care rates for Human Immunodeficiency Virus (HIV) \cite{khatami2021reinforcement}. They define discrete actions for the HIV testing and AntiRetroviral Therapy (ART) using the proxy indicators of decrement in the unaware population rate ($\{0\%, -2.5\%\}$) and increment in the ART proportion (\{0\%, 10\%, 20\%\}) for two kinds of risk population, i.e., HETerosexuals (HETs) and Men who have Sex with Men (MSM).
Kwak et al. adopted Dueling Double Deep Q-Network (D3QN) to optimize the joint policy of domestic lockdown and travel restrictions \cite{kwak2021deep}. They divide the intensity of those two types of intervention into three discrete levels, respectively, to construct a $3 \times 3$ action space: (1) no action, restricted public social gathering, and nationwide lockdown for domestic lockdown policy; and (2) no action, flight suspension, and full closure of all borders for travel restrictions.

Some works adopt the policy gradient methods to deal with the continuous action space.
For instance, Ghazizadeh et al. apply the Deep Deterministic Policy Gradient (DDPG) to optimize the joint strategy of three kinds of control interventions: quarantine of susceptible people, vaccination of the susceptible group, and treatment of symptomatic infected individuals, which are formulated as continuous values ranging from 0 to 1 \cite{ghazizadeh2024modeling}.
Kao et al. optimized two key interventions daily implemented for COVID-19: constraint of moving distances and screening area scope \cite{kao2024dynamic}. They construct the meta-population Susceptible-Exposed-Infected-Quarantined-Removed (SEIQR) model comprising four regions and further set the circle range with a radius of $50\sqrt{2}$ pixels for each region and transport hubs with a 5-pixel radius centered in the regions. Then, they define the control value of moving distances and screening area in a continuous domain ranging from 1 to 5 pixels and from 0 to 10 pixels, respectively, and learn optimal values by the Asynchronous Advantage Actor-Critic (A3C) algorithm.
Mannarini et al. aimed to find the optimal mix policy of nine interventions at the country level with the DDPG algorithm, making the trade-off between the viral transmission, indicated by the effective reproduction number, and the economic cost, indicated by the unemployment rate \cite{mannarini2022if}. They consider 9 types of non-pharmaceutical interventions, listed in Table \ref{tab: multi-interventions}. For each intervention, there are four discrete and ascending stringency levels. Due to the huge exploration space for the joint action caused by the multiple interventions, they formulate the output of the policy network for each measure as a continuous value between -1 and 1 learned by the policy-gradient method, to indicate the extent of increase or decrease in stringency levels from the previous policy. Furthermore, firstly, they constrain value scopes according to the real policy dataset and set a weekly frequency of policy change, and then convert the continuous value of the policy to the discrete levels by rounding.
Bushaj et al. proposed the Simulation-Deep Reinforcement Learning (SiRL) model to optimize the deployment of multiple interventions \cite{bushaj2023simulation}. Different from learning the types and intensity of interventions in the combination strategy via the RL method, they manually define nine actions that are composed of various types of common interventions, which are presented in Table \ref{tab: multi-interventions}, for the Deep Reinforcement Learning (DRL) agent to select from. This design can significantly reduce the solution space, but it could rely on the empirical knowledge from public health experts. Their algorithm also formulates a multi-objective reward function to find the trade-off between economic stability and well-being at the same time.

Besides, some works develop novel algorithms to improve the effectiveness and efficiency of multiple interventions learning. For instance, Du et al. proposed a novel RL decision framework, named Hierarchical Reinforcement Learning for Epidemic Control (HRL4EC), to address the multi-mode disease control \cite{du2023hrl4ec}. Exemplifying with three interventions for COVID-19, i.e., mobility constraint, temporary medical resources, and necessities supply, this algorithm develops a hierarchical decision process based on the PPO model. The disease decision-making is decomposed into two stages to reduce the complexity of the solution space: at the high level, the model determines which types of interventions will be deployed; at the low level, the model determines when and how to deploy them.

Most studies reviewed in this section directly adopt the standard RL algorithms to learn combining strategies of multiple interventions. A common simplification in these works is the use of small, discrete action spaces, which are typically predefined as a limited set of intervention levels. However, along with the increase in considered control measures and the finer granularity in resolution of their available value range, sample or computational efficiency needs to be considered to ensure algorithms can reliably converge to robust and effective policies in the resulting high-dimensional action space.
However, few algorithms explore the design of mechanisms to reduce the solution space and improve the efficiency of the learning process, such as the hierarchical framework and incorporating expert knowledge, posing a potential research direction for future research.

\section{Inter-regional Coordinated Control}\label{sec: Inter-regional}
Due to intensive travel flows facilitated by the highly developed transportation networks, such as subway lines and global airport networks, the rapid and widespread propagation of infectious diseases, specifically respiratory infectious diseases \cite{wilder2021covid}, is becoming a critical global health risk. The lack of coordinated policy strategies of different administrative regions responding to disease outbreaks will aggravate threats of disease transmission in both individual and collective regions' perspectives \cite{jit2021multi}.
However, stemming from the gaps in information sharing, anti-epidemic resources, and travel restrictions, the inter-regional coordinated control of infectious diseases is difficult and remains a serious challenge.
These disparities often lead to conflicting policies resulting in misaligned efforts, where one region's containment strategy can be undermined by a neighboring region's inaction, so as to significantly reduce the overall effectiveness of measures.
To tackle this problem, RL methods have been adopted to explore coordinated policy among multiple administrative regions.

Research on the use of RL for inter-regional coordinated control of infectious diseases remains limited.
The scarcity of literature is partly due to the significant real-world implementation challenges, such as incompatible data systems across jurisdictions, conflicting interests of multiple stakeholders, and privacy issues of information.
Within the literature search scope outlined in Section \ref{sec: Method}, there is only one paper included. Specifically, Khatami and Gopalappa constructed a Susceptible-Exposed-Infected-Recovered-Dead (SEIRD) model with human mobility between two geographical jurisdictions and explored the effectiveness of lockdown policy (defined as the four discrete levels) under the cases of cooperation and non-cooperation of them \cite{khatami2022deep}. To evaluate the effects of the cooperative and the uncooperative policy of two jurisdictions, denoted as A and B, they defined the uncooperative situation between A and B with A adopting the optimal policy and B having no interventions, while the cooperative situation is that B sets the exact same policy as region A, which is based on the assumption that A and B emerge the disease transmission at the same time. Furthermore, in a series of different preset scenarios, they trained the Deep Q-Networks (DQN) to learn the optimal policy of A, based on the designed objective with two components (economic burden and hospital capacity) and three components (aforementioned two plus the number of deaths), respectively. In fact, they only use the single RL model to learn the optimal policy for jurisdiction A.
It could oversimplify the scenario of the policy learning of different regions, which did not take the instability caused by the interactions of policies and epidemic dynamics among different regions into consideration.

Actually, exceeding the searching scope of this survey, there are some works that formulate the inter-regional coordinated control of infectious disease spread under the paradigm of Multi-Agent Reinforcement Learning (MARL) \cite{luo2025architecting, luo2025h2}, where each region or country is considered as an agent to adaptively and autonomously choose its own policy while taking the real-time epidemic status and deployed policies of other agents into account. However, even in recent years, studies focusing on this promising area have been scarce, suggesting that it could be a potential direction for future research.

\section{Conclusions and Future Directions}\label{sec: Future}
In this survey, we explore the utility of RL methods in infectious disease control by reviewing and discussing a series of the latest literature, specifically in how the RL methods assist decision-making via adaptive optimization of interventions. Furthermore, we categorize the existing works of this field into four topics related to public health needs distinguished by the problem definition and formulation: (1) resource allocation; (2) balancing between lives and livelihoods; (3) mixed policy of multiple interventions; and (4) inter-regional coordinated control. For each topic, we comb and summarize a branch of works addressing these challenges.

Based on the previous discussion of existing literature within the search scope, we further identify the following potential directions to spark future research to achieve more intelligent and adaptive decision-making for infectious disease control:
\begin{itemize}
    \item \textbf{Effectively learning intervention combinations in the large solution space.}
    In settings of a wide variety of intervention types with continuous values, there is a pressing need for efficient algorithms for learning optimal intervention combinations because of the induced enormous state-action space.
    Existing studies have employed classical value-based methods to learn optimal combinations of several control measures with discrete action values \cite{kwak2021deep}, and some studies adopted the policy gradient method to explore the continuous action space \cite{ghazizadeh2024modeling}. In addition, hierarchical decision-making frameworks have been proposed to learn the mask to reduce the solution space \cite{du2023hrl4ec}. Nevertheless, this problem remains underexplored in the background of infectious disease control. Further research is necessary to extend these approaches to large-scale problems by designing mechanisms or integrating empirical knowledge to learn robust and effective solutions.
    
    \item \textbf{Enhancing coordination between multiple administrative regions.}
    In the background of multiple regions working together to fight against the rapidly spreading epidemic, the learning of a coordinated joint strategy becomes more and more crucial due to the mutual effect between multiple regions' policies.
    MARL provides a powerful framework to formulate problems and discover the solutions in this complex case, where each administrative region can be recognized as an agent that makes decisions independently or collaboratively \cite{luo2025h2}. However, from the algorithm perspective, optimizing a joint policy for multiple regions remains challenging due to the vast solution space, the non-stationarity induced by the dynamical environment, and partial observability \cite{nguyen2020deep}. Furthermore, the governments are not always in fully cooperative or fully competitive scenarios with each other due to the scarce total resources and the consideration for the region's own interests. It illustrates that multiple objectives and conflicting interests should also be considered in the algorithm designs. However, so far, RL or MARL methods exploring those challenges of coordinated epidemic control are still limited. Existing algorithmic innovations in MARL algorithms have explored these aspects in the context of games \cite{wang2022cooperative}. In future works, how to utilize and adopt these experiences to address these challenges in the context of infectious disease control is one of the frontiers for artificial intelligence assistant public health policy.
    
    \item \textbf{Constructing standard and general benchmarks for uniform comparison of diverse algorithms and indicators.}
    First of all, most existing work has constructed different types of infectious disease transmission environments for the epidemic and intervention simulation to fit their study problems in terms of the targeted diseases and intervention types. For example, the agent-based transmission modeling via social networks and meta-population disease modeling based on the regional adjacency can be used to emulate vaccination strategy at the individual level \cite{dong2024integrating} and population level \cite{ling2024cooperating}.
    In addition, these simulation environments typically consist of different compartments, various parametric settings, network configurations, and intervention capacity, and are calibrated with datasets of different time durations and regions. The RL methods are usually adapted to those settings to derive the model structure designs.
    Although those works can provide a perspective for addressing the fixed problems under a specific epidemic context, it is difficult to compare the efficiency performance with regard to various intervention types, RL method types, or diverse health risk and economic indicators across the context directly.
    Therefore, how to build and construct standard and general benchmarks for uniform comparison of diverse algorithms and indicators is a promising direction for the development of innovative algorithms.
\end{itemize}

\section*{Acknowledgment}

This work was supported in part by the National Science and Technology Major Project under Grant No. 2021ZD0112500, the General Research Fund from the Research Grant Council of Hong Kong SAR under Projects RGC/HKBU12203122 and RGC/HKBU12200124, the NSFC/RGC Joint Research Scheme under Project N\_HKBU222/22, and the Guangdong Basic and Applied Basic Research Foundation under Project 2024A1515011837.

\bibliographystyle{IEEEtran}
\bibliography{IEEEabrv,ref}

\end{document}